\documentclass{optica-article}

\usepackage{opticajournal}

\usepackage{lineno}
\usepackage{subfigure}
\usepackage{amsmath}
\usepackage{soul}

\begin{document}

\title{Asynchronous Event Stream Noise Filtering for High-frequency Structure Deformation Measurement}

\author{Yifei Bian\authormark{1,2}, Banglei Guan\authormark{1,2,*}, Zibin Liu\authormark{1,2}, Ang Su\authormark{1,2}, Shiyao Zhu\authormark{1}, Yang Shang\authormark{1,2} and Qifeng Yu\authormark{1,2}}

\address{\authormark{1} College of Aerospace Science and Engineering, National University of Defense Technology, Changsha 410073, China \\
\authormark{2} Hunan Provincial Key Laboratory of Image Measurement and Vision Navigation, Changsha 410073, China}

\email{\authormark{*}guanbanglei12@nudt.edu.cn} 

\begin{abstract} 
Large-scale structures suffer high-frequency deformations due to complex loads. However, harsh lighting conditions and high equipment costs limit measurement methods based on traditional high-speed cameras. This paper proposes a method to measure high-frequency deformations by exploiting an event camera and LED markers. Firstly, observation noise is filtered based on the characteristics of the event stream generated by LED markers blinking and spatiotemporal correlation. Then, LED markers are extracted from the event stream after differentiating between motion-induced events and events from LED blinking, which enables the extraction of high-speed moving LED markers. Ultimately, high-frequency planar deformations are measured by a monocular event camera. Experimental results confirm the accuracy of our method in measuring high-frequency planar deformations.
\end{abstract}

\section{Introduction}
Large-scale structures carry complex loads, resulting in high-frequency deformations within their two-dimensional plane~\cite{Shien}. Timely measurement of these structure deformations is crucial for health monitoring and operation.

Such structures typically span hundreds of meters, necessitating deformation measurement accuracy at key nodes to achieve the millimeter level. Various sensors and methods are currently utilized for measuring structure planar deformations, including accelerometers, strain gauges, fiber Bragg gratings, and Doppler LiDAR~\cite{Jin,Yi,Li-Zhuang}. Accelerometers are effective for measuring structure deformations such as steel frame bridges due to their high sampling rate. However, accelerometers require double integration, making them unsuitable for low-frequency deformation monitoring~\cite{Moschas,Nan}. Strain gauges enable convenient monitoring across multiple locations at a reduced cost. However, strain gauges require pre-selection of gauge positions to detect strain at specific points. Furthermore, over extended monitoring periods, these gauges are susceptible to failure due to factors such as creep and temperature variations~\cite{Bezziccheri}. Fiber Bragg Grating (FBG) embedding within the structure may impact its mechanical properties~\cite{Ziyue}. Doppler LiDAR functions is a non-contact measurement method. Nonetheless, the Doppler LiDAR is expensive and produces massive point cloud data, necessitating substantial computational 
resources~\cite{Beshr,Miskiewicz}.

Videogrammetry is also utilized for planar structure deformation measurement~\cite{Lin,Dongming,James}. Henke~\cite{Henke} et al. utilized LED markers and digital image processing technology to measure structure deformations. Traditional frame-based cameras are influenced significantly by challenging lighting conditions. In dark environments, detecting LED markers may lead to overexposure, thereby impacting measurement accuracy. Zhang~\cite{Zhang} et al. proposed a method for monitoring bridge deformations and analyzing dynamic characteristics utilizing a monocular camera. However, the high-speed camera generates massive images, presenting challenges for data processing. In summary, the existing videogrammetry method for measuring high-frequency planar structure deformation is inadequately suited for harsh lighting conditions. Additionally, the costs associated with hardware and data processing for measuring high-frequency deformations are prohibitively high.
\begin{figure}[htbp]
\centering\includegraphics[width=12.5cm]{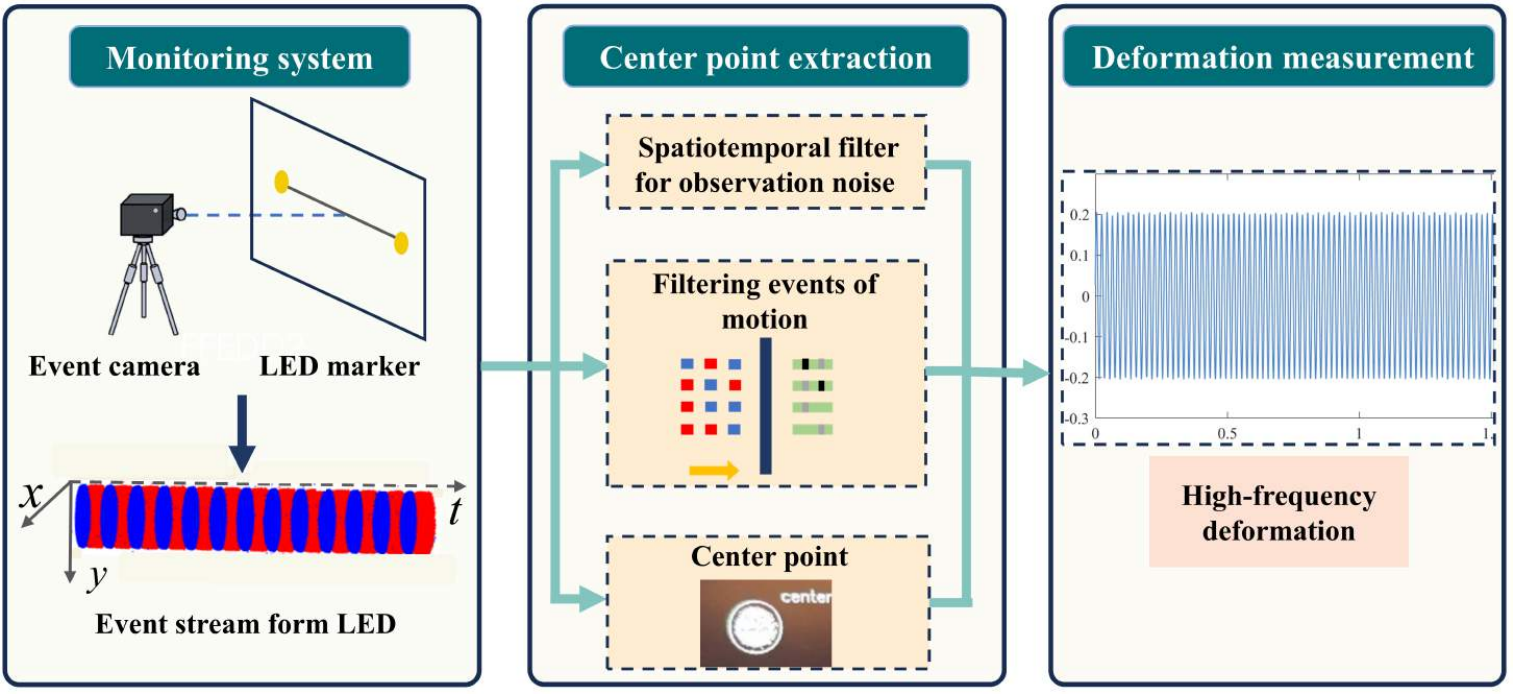}
\caption{Flow diagram of the proposed method.}
\label{fig1}
\end{figure}

Therefore, this paper proposes a method to measure high-frequency planar deformations utilizing the event camera, as illustrated in Fig. \ref{fig1}. Event cameras have advantages like high temporal resolution, high dynamic range, low power consumption, and low latency. These capabilities mitigate challenges that traditional frame-based cameras struggle with. The paper utilizes an event camera and LED markers to measure structure deformations in a plane. Firstly, the observation noise is reduced based on spatiotemporal correlation, which improves the efficiency of marker extraction. Moreover, frequency analysis of event polarity reversals distinguishes events generated by marker motion from those caused by blinking. In addition, the accurate extraction of pixel coordinates for the center points of LED markers is accomplished via the asynchronous event stream. Ultimately, the LED markers are extracted, enabling the deformation measurements by the monocular event camera. The proposed method achieves accurate measurement of high-frequency planar deformation, offering low cost and improved efficiency. The innovative aspects of our method are as follows:
\begin{itemize}
    \item We leverage the characteristics of the event stream generated by LED markers blinking and employ spatiotemporal correlation to filter observation noise, thereby attaining better denoising outcomes.
    \item We extract pixel coordinates for the center points of LED markers from the asynchronous event stream, enabling the extraction of high-speed moving LED markers.
    \item We utilize an event camera and LED markers to measure the high-frequency structure deformation, offering a cost-effective and adaptable solution.
\end{itemize}

The paper is structured as follows: Section 2 reviews the related work. Section 3 introduces the method, including reducing observation noise, extracting center points of LED markers, and measuring deformation. Section 4 presents experimental results to validate the proposed methods. Section 5 offers a conclusion.

\section{Related work}
The event camera is an innovative image sensor inspired by biological vision, generating events asynchronously in response to pixel-level light intensity changes~\cite{Guillermo}. Event cameras detect the brightness change in the field-of-view of the respective pixel, and output events asynchronously as the change exceeds a threshold. Event cameras possess high temporal resolution, high dynamic range, low latency, and low power consumption, enhancing cost-effectiveness compared to traditional high-speed cameras. Previous studies employing event cameras for structural deformation measurements have demonstrated good results. Zhu~\cite{QZhang} et al. integrated an event camera with digital scattering technology to observe stroboscopic flash scattering. They subsequently employed Digital Image Correlation (DIC) to capture structure strain, thereby achieving high-frequency deformation measurement. This approach necessitates scattering material on the surface, where the quality of scattering directly affects measurement accuracy. Lai~\cite{Lai} et al. employed an event camera to measure structure deformations. They introduced an innovative sparse identification framework based on physical information for vibration monitoring and mechanical analysis of beams. This method operates without markers, relying exclusively on the event camera. However, its application in low-light conditions remains limited.

The output of the event camera includes observation noise that could impact the accuracy of extracting LED markers center. There are various methods available for reducing noise. Delbruck~\cite{delbruck2008frame} introduced a method that reduces noise by recording the temporal and spatial occurrence of an event and comparing it with subsequent events in the same area. This method necessitates individual evaluation for each event, leading to inefficiencies that might mistakenly filter valid outputs at the beginning of object motion. Li~\cite{Mengjie} et al. introduced content correlation of event polarity changes leveraging spatiotemporal correlation, modeling light intensity change caused by moving objects to preserve image edges. In this paper, we propose a denoising approach based on a spatiotemporal correlation suit for event streams generated by LED markers blinking, demonstrating more efficiency.

The commonly used method for tracking LED markers with event cameras typically neglects the influence of motion-induced events on cluster tracking~\cite{Litzenberger,Litzenberger1}. Lagorce~\cite{Lagorce} et al. employed Gaussian kernel asynchronous iteration to track features based on incoming events, integrating spatial and temporal correlations. Various kernel functions can also be utilized to track targets with diverse shapes. Nevertheless, the effect of target motion on the events is not explicitly addressed. Censi~\cite{Andrea} et al. mounted LED markers on an unmanned aerial vehicle and proposed a low-latency tracking method that correlates LED markers based on the time interval between events, and then tracks each LED using a particle filter.

\section{Methods}

\subsection{Spatiotemporal correlation based denoising}
To accurately extract the coordinates of the center points for LED markers, the first step is to reduce observation noise. The main observation noises are background noise and thermal noise. Background noise lacks correlation with surrounding events, while thermal noise is similarly uncorrelated with other events in space. This paper introduces a novel denoising method that leverages spatiotemporal correlation, designed specifically for the characteristics of events generated by LED markers blinking. Define an event using Eq. (\ref{eq1}):

\begin{equation}
e=[\mathbf{x},t,s],
\label{eq1}
\end{equation}
where $e$ presents an event, $\mathbf{x}$ is the pixel coordinate of the event, $t$ is the timestamp, and $s$ indicates the polarity. The event polarity is the sign of the brightness change. When $s=1$, it indicates an increase in brightness; conversely, $s=0$ indicates a decrease in brightness.
\begin{figure}[htbp]
\centering\includegraphics[width=8cm]{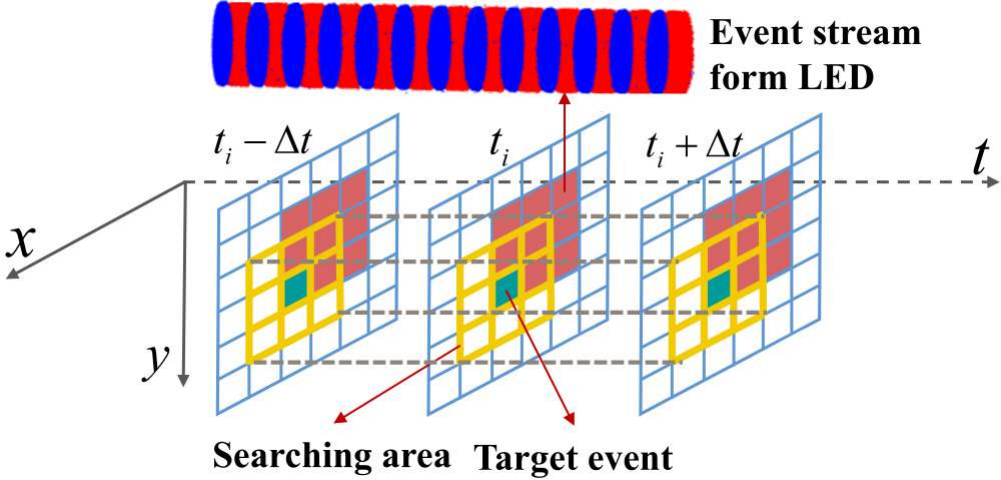}
\caption{Spatiotemporal correlation-based denoising. Both the space adjacent to the target event and the separated time $\Delta t$ from the timestamp of the target event are searching areas.}
\label{fig2}
\end{figure}

As depicted in Fig. \ref{fig2}, the event stream produced by LED markers exhibits concentration centrally within a region. These events share identical timestamps and display a highly spatial correlation, with numerous events closely grouped together. If the target event resides within the event cluster, it must be associated with spatially correlated events.

Firstly, assess the number of events under the same timestamps to filter out those not belong to the LED light-generated event cluster. As demonstrated in Eq. (\ref{eq2}):
\begin{equation}
{{e}_{n}}=\left\{ [\mathbf{x},{{t}_{i}},s]|{{N}_{i}}=\sum\limits_{{{t}_{i}}}{[\mathbf{x},{{t}_{i}},s],}\text{ }{{N}_{i}}<{{N}_{\text{th}}} \right\},
\label{eq2}
\end{equation}
where $e_{n}$ means noise events. $N_{i}$ is the number of all events at the timestamp $t_{i}$. $N_{\text{th}}$ is a specified threshold for the number of events occurring at the same timestamp, determined by the size and blinking frequency of the LED markers. The value of $ N_{\text{th}}$ should be larger when the size of the marker is larger and the blinking frequency is higher. If $N_{i}\le N_{\text{th}}$, it can be assumed that none of the events at that moment are generated by the LED blinking and should be filtered out as noise events.

The coarsely filtered events undergo further refinement by evaluating the presence of related events in their spatiotemporal correlation. Assume the target event ${{e}_{i}}=[{{x}_{i}},{{y}_{i}},{{t}_{i}},s]$, $T_{x}$ and $T_{y}$ represent the spatial distance threshold between the two events, respectively. $T_{t}$ denotes the threshold value associated with the event in the time domain. If there is an event ${{e}_{j}}=[{{x}_{j}},{{y}_{j}},{{t}_{j}},s]$ and satisfies Eq. (\ref{eq3}):
\begin{equation}
[{{x}_{j}}\in U\left( {{x}_{i}},{{T}_{x}} \right)]\text{ }\wedge \text{ }[{{y}_{j}}\in U\left( {{y}_{i}},{{T}_{y}} \right)]\text{ }\wedge \text{ }[{{t}_{j}}\in U\left( {{t}_{i}},{{T}_{t}} \right)],
\label{eq3}
\end{equation}
In that case, the target event $e_{i}$ is determined not to be observation noise. $(T_{x},T_{y},T_{t})$ is determined by the scenario of event output. For the event output scenario of LED markers blinking in this paper, these values should be small.

\subsection{Reduce motion generation events by polarity}
Besides events caused by blinking, LED markers also trigger events due to motion. These events may obscure the edges of the LED markers and affect marker extraction. The events generated by LED marker motion and blinking are closely associated temporally and spatially. Therefore, reducing motion-generated events using spatiotemporal correlation is ineffective.

\begin{figure}[htbp]
\centering\includegraphics[width=8cm]{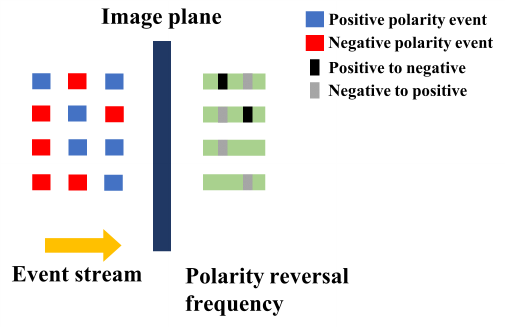}
\caption{Reduce motion generation events based on polarity reversal frequency. Record the frequency of transition from positive to negative polarity and negative to positive polarity.}
\label{fig3}
\end{figure}

The LED markers blinking produces light intensity changes in the field-of-view of the event camera, resulting in events with different polarities. As is seen in Fig. \ref{fig3}, the LED markers blink at a modulated frequency, then record the type and frequency of event polarity reversal. Let ${{s}_{o\_ f}}({{x}_{i}},{{y}_{i}})$ denote a reversal from a positive to a negative polarity event at pixel location $({{x}_{i}},{{y}_{i}})$ and ${{s}_{f\_ o}}({{x}_{i}},{{y}_{i}})$ denote a shift from a negative to a positive polarity event at that location, Eq. (\ref{eq4}) gives the relationship between the polarity reversal frequency and the LED blinking frequency:
\begin{equation}
{\frac{1}{\Delta t(s)}=\frac{1}{t\left[ s_{o\_f}^{1}({{x}_{i}},{{y}_{i}})-s_{o\_f}^{2}({{x}_{i}},{{y}_{i}}) \right]}=\frac{1}{t\left[ s_{f\_o}^{1}({{x}_{i}},{{y}_{i}})-s_{f\_o}^{2}({{x}_{i}},{{y}_{i}}) \right]}={{f}_{i}}},
\label{eq4}
\end{equation}
where $t\left[  \right]$ denotes the time interval between two polarity reversals. For clusters of events generated by LED blinking, the frequency polarity reversal aligns with the blinking. Events generated by motion exhibit a low frequency of polarity reversal, and a significant disparity compared to events generated by LED blinking. This allows for the determination of whether the events generated by LED markers, as Eq. (\ref{eq5})
\begin{equation}
e=\left\{\begin{array}{l}{\left[x_{i}, y_{i}, 1, t\right]} \\{\left[x_{i}, y_{i}, 0, t\right]} \\{\left[x_{i}, y_{i}, 1, t+\Delta t(s)\right]} \\{\left[x_{i}, y_{i}, s, 0+\Delta t(s)\right]}\end{array} \text { noise, if } \frac{1}{\Delta t(s)} \notin U\left(f_{i}-f_{\text{TH}}\right)\right. \text {. }
\label{eq5}
\end{equation}
where $f_{\text{TH}}$ represents the threshold for frequency, $U(f_{i}-f_{\text{TH}})$ denotes the interval identified as the frequency of LED blinking, and $e$ represents the four events separating two polarity reversals. If the frequency of polarity reversal does not align with that of the LED markers, it is presumed that $e$ originates from the marker motion and needs to be reduced.

\subsection{Event-based marker center tracking}
The distribution of the event clusters $E$ in the spatiotemporal domain can be approximately modeled as a two-dimensional Gaussian distribution $N(\mu,\Sigma)$, characterized by the following probability density function as Eq. (\ref{eq6}):
\begin{equation}
f(\mathbf{x})=\frac{1}{2\pi \centerdot {{\left| \Sigma  \right|}^{{1}/{2}\;}}}\exp \left[ -{1}/{2}\;{{(\mathbf{x}-\mu )}^{T}}{{\Sigma }^{-1}}(\mathbf{x}-\mu ) \right],\mathbf{x}=(x,y),
\label{eq6}
\end{equation}
where the Gaussian mean $\mu$ describes the location of the event clusters and can be replaced with its estimated value as Eq. (\ref{eq7}):
\begin{equation}
\mu =\overline{\mathbf{x}}=\frac{1}{n}\sum\limits_{i=1}^{n}{{{\mathbf{x}}_{i}}},
\label{eq7}
\end{equation}
where $n$ denotes the total number of events within a cluster. The covariance matrix is a second-order square matrix that characterizes the shape and size of the event cluster, expressed as Eq. (\ref{eq8}):
\begin{equation}
\Sigma =\left[ \begin{matrix}
   \sigma _{x}^{2} & {{\sigma }_{xy}}  \\
   {{\sigma }_{yx}} & \sigma _{y}^{2}  \\
\end{matrix} \right]
\label{eq8}
\end{equation}

The optical axis is perpendicular to LED markers during photographed, and a standard circular shape is obtained, so ${{\sigma }_{xy}}={{\sigma }_{yx}}=0$.

For each newly generated event $e'=[\mathbf{x}',t',s]$, compute the Mahalanobis distance between the cluster of events $E$ and $e$:
\begin{equation}
d(\mathbf{x}',E)=\sqrt{{{(\mathbf{x}'-\mu )}^{T}}{{\Sigma }^{-1}}(\mathbf{x}'-\mu )},
\label{eq9}
\end{equation}
where $\mathbf{x}'$ denotes the pixel coordinates of the new event, $d(\mathbf{x}',E)$ represents the Mahalanobis distance from the event to the overall event cluster $E$, and ${{d}_{\text{TH}}}$ is the threshold for determining whether an event belongs to an event cluster or not. Then:
\begin{equation}
\mathbf{x}\in E,\text{if} \ \ d(\mathbf{x},E)<{{d}_{\text{TH}}},
\label{eq10}
\end{equation}
If the new event belongs to this event cluster, refer to Eq. (\ref{eq7}) and update the position of the event cluster.

For an event $e''=[\mathbf{x}'',t'',s]$ that has already been added to the event cluster, it is necessary to eliminate the impact of this event promptly. Let ${{t}_{\text{new}}}$ denote the timestamp of the most recent event, and the time that an event can exist in the event cluster be ${{t}_{\text{su}}}$. When $t''-{{t}_{\text{new}}}>{{t}_{\text{su}}}$, the events that should not have been added to the event cluster had existed for too long. Remove $e''$ and reference Eq. (\ref{eq7}) to update the position of the event cluster accordingly.

\subsection{Planar deformation measurement}
\begin{figure}[htbp]
\centering\includegraphics[width=8cm]{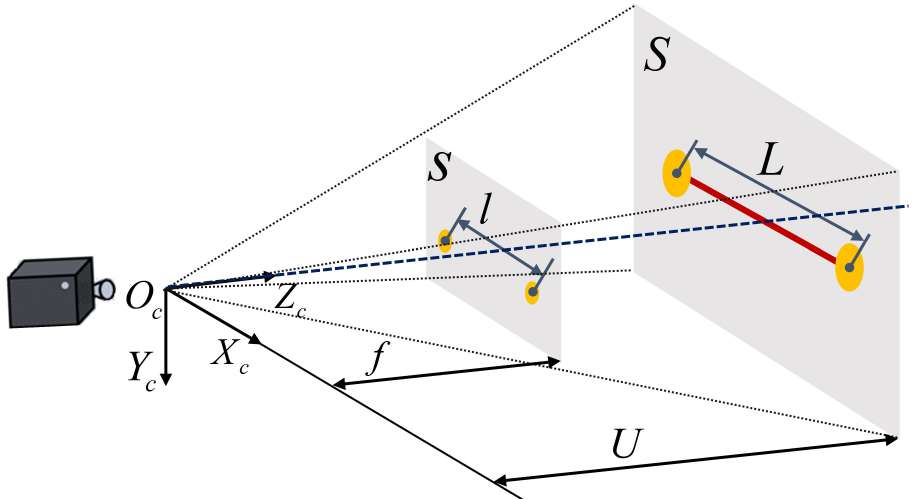}
\caption{Planar deformation measurement employing monocular vision. ${{O}_{c}}-{{X}_{c}}{{Y}_{c}}{{Z}_{c}}$ represents the camera coordinate system. $s$ is the image plane and $S$ is the plane to be measured. ${{O}_{c}}-{{X}_{c}}$ axis is perpendicular to the $S$.}
\label{fig4}
\end{figure}

When the optical axis of the event camera is perpendicular to the structure, a similarity relationship dictated by the principle of central projection is established. The markers of known dimensions are affixed to the structure to calculate the magnification for planar measurements. In this study, two LED markers are interconnected by rigid rods of predetermined length, as illustrated in Fig. \ref{fig4}. The center points of LED fixation are settled previously. 

Let $O_{c}$ denote the optical center, with the coordinate system of the camera represented as ${{O}_{c}}-{{X}_{c}}{{Y}_{c}}{{Z}_{c}}$. $s$ is the image plane and $S$ is the plane to be measured. $U$ represents the object distance, defined as the distance from the optical center to the object. $f$ signifies the focal length, which is the distance from the optical center to the image plane. When the $Z_{c}$-axis is perpendicular to the plane to be measured:

\begin{equation}
\frac{L}{l}=\frac{U}{f},
\label{eq11}
\end{equation}
where $L$ represents the distance between the fixed center positions of two LED markers, while $l$ denotes the distance between the center points in the image plane.

When the structure deforms at high frequency, the rigid rod ensures that the distance between the LED markers remains constant. Let the displacements of the markers in the image coordinate system be $\Delta u$ and $\Delta v$, respectively. The actual displacement in the camera coordinate system can be calculated by Eq. (\ref{eq12}):

\begin{equation}
\Delta X=\frac{L}{l}\centerdot \Delta u,\text{ }\Delta Y=\frac{L}{l}\centerdot \Delta v,
\label{eq12}
\end{equation}
where $\Delta X$ represents the displacement of the structure in the ${{O}_{c}}-{{X}_{c}}$-axis direction and $\Delta Y$ represents the displacement of the structure in the ${{O}_{c}}-{{Y}_{c}}$-axis direction.

\section{Experiments}

\subsection{Observation noise suppression}
\begin{figure}[ht]
\centering
\subfigure[LED marker that can control blinking frequency.]{   
\begin{minipage}{0.5\linewidth}
\centering 
\includegraphics[height = 4.5cm]{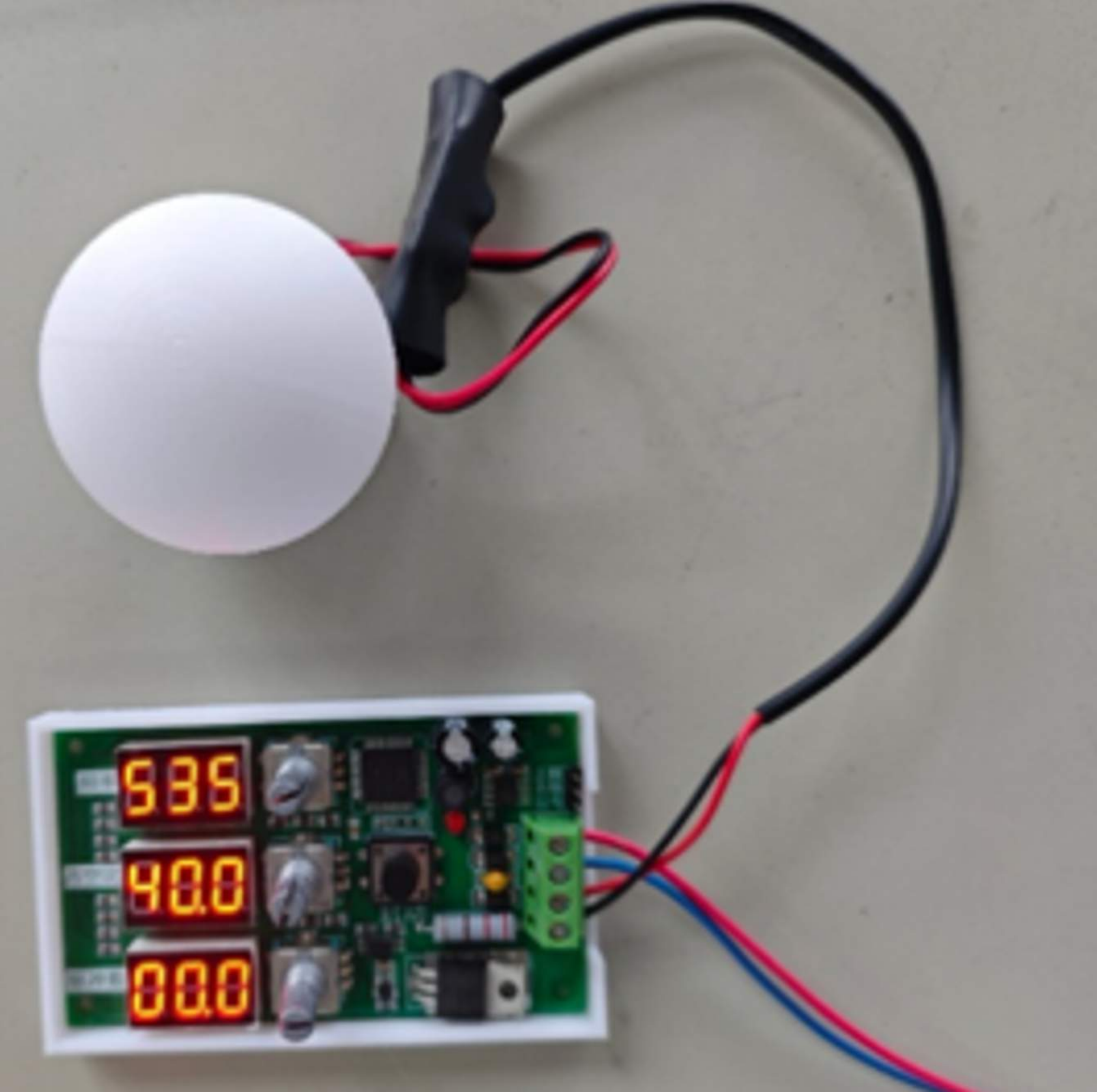}  
\end{minipage}
\label{fig5a}
}\subfigure[Event camera observing LED markers blinking.]{ 
\begin{minipage}{0.5\linewidth}
\centering  
\includegraphics[height = 4.5cm]{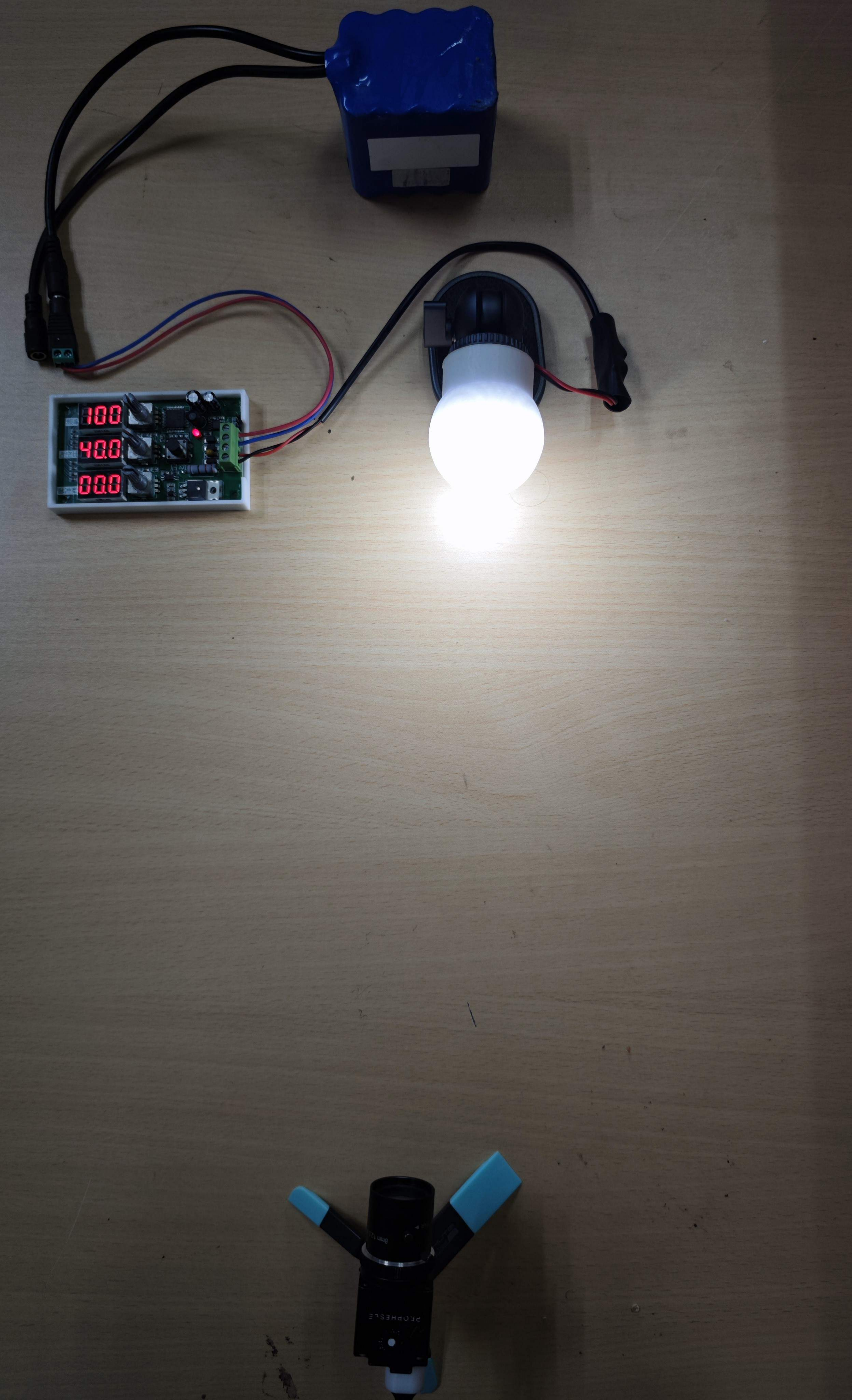}
\end{minipage}
\label{fig5b}
}
\caption{Configuration of the observation noise suppression experiment.}   
\label{fig5}   
\end{figure}

\begin{figure}[htbp]
\centering\includegraphics[width=8cm]{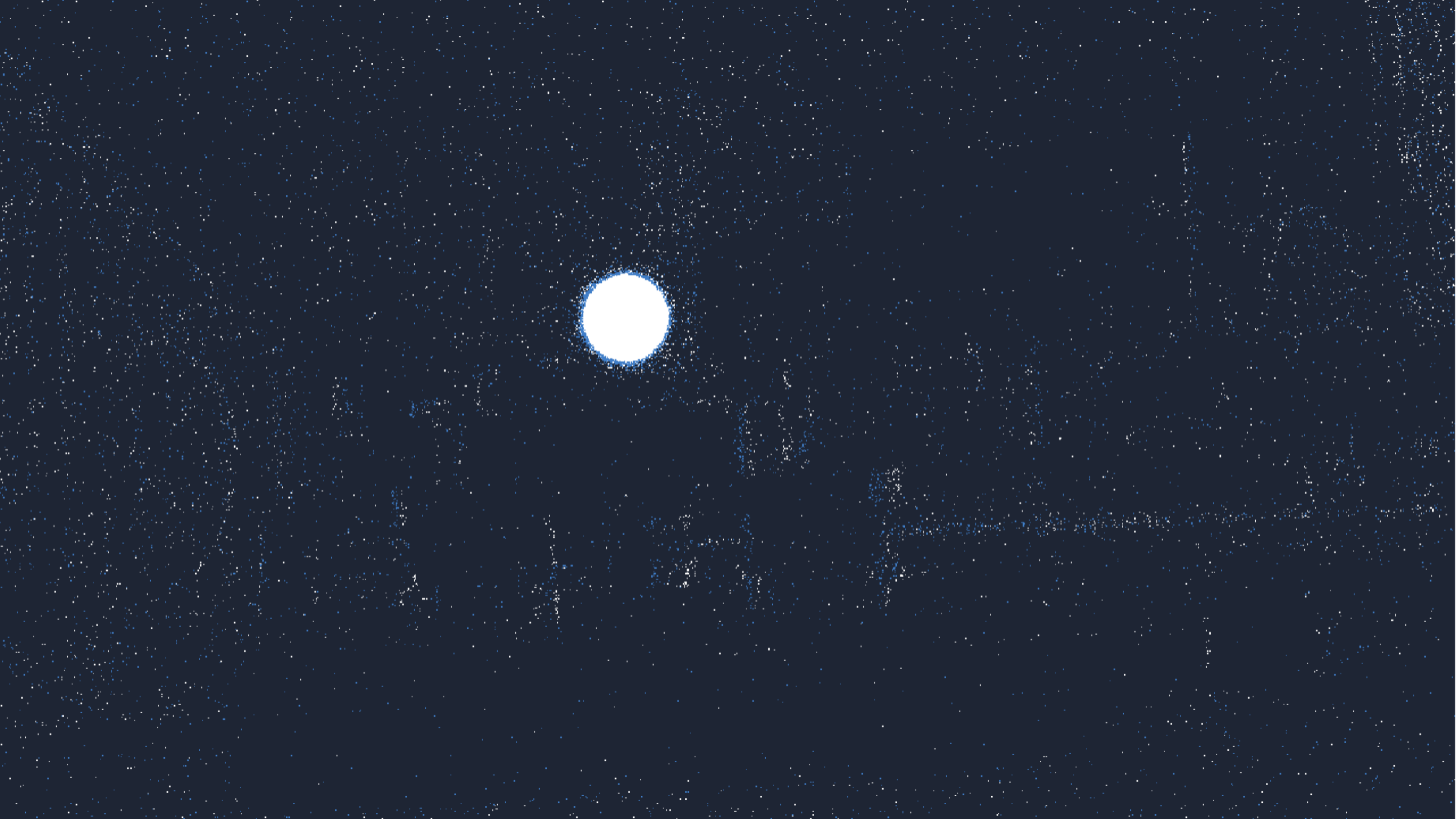}
\caption{Image within the camera's field of view.}
\label{fig6}
\end{figure}

An experiment was performed to validate the efficacy of the denoising method proposed in this paper. Fig. \ref{fig5} depicts the experimental configuration. The LED marker is shown in Fig. \ref{fig5a}. Set the blinking frequency to{$100\enspace \text{Hz}$, and the light to dark time ratio in one blink to $2:3$. The event camera utilized in the experiment is Prophesee EVK4 (resolution $1280\times 720$ pixels, dynamic range $86\enspace \text{dB}$, and latency $220\enspace \mu \text{s}$). The camera is equipped with a lens with a focal length of $12\enspace \text{mm}$, and the camera is $50\enspace \text{cm}$ from the marker, as shown in Fig. \ref{fig5b}. The scene in the camera's field of view is depicted in Fig. \ref{fig6}. Data are captured for 2.96 seconds during the experiment. 

Fig. \ref{fig7a} illustrates the raw output of the event camera containing massive noise events. The observation noise in the spatiotemporal domain could potentially compromise the accurate extraction of the marker center point. Fig. \ref{fig7b} presents a coarse filtering of the raw event stream data. The minimum number of events at the identical timestamp is set as ${{N}_{th}}=5$. The observation noise noticeably decreases after applying coarse filtering. Fig. \ref{fig7c} illustrates noise reduction by our method. This filtering method effectively eliminates observation noise. Fig. \ref{fig7d} applies noise filtering to the raw event stream exclusively based on spatiotemporal correlation. It is evident that the noise persists after denoising.

\begin{figure}[h]
\centering
\subfigure[Raw stream event.]{   
\begin{minipage}{0.5\linewidth}
\centering 
\includegraphics[height = 4.5cm]{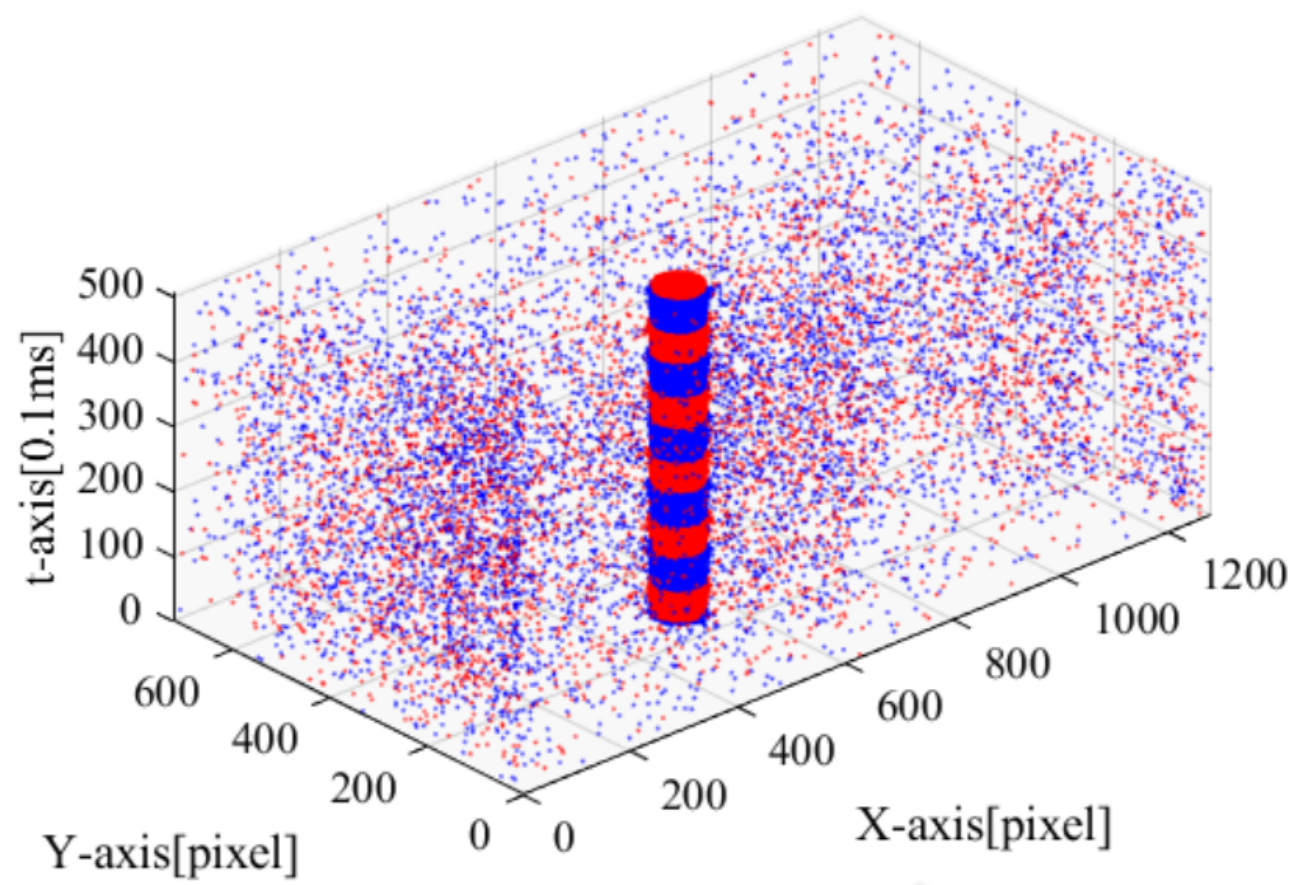}  
\end{minipage}
\label{fig7a}
}\subfigure[Coarsely filtered events.]{ 
\begin{minipage}{0.5\linewidth}
\centering  
\includegraphics[height = 4.5cm]{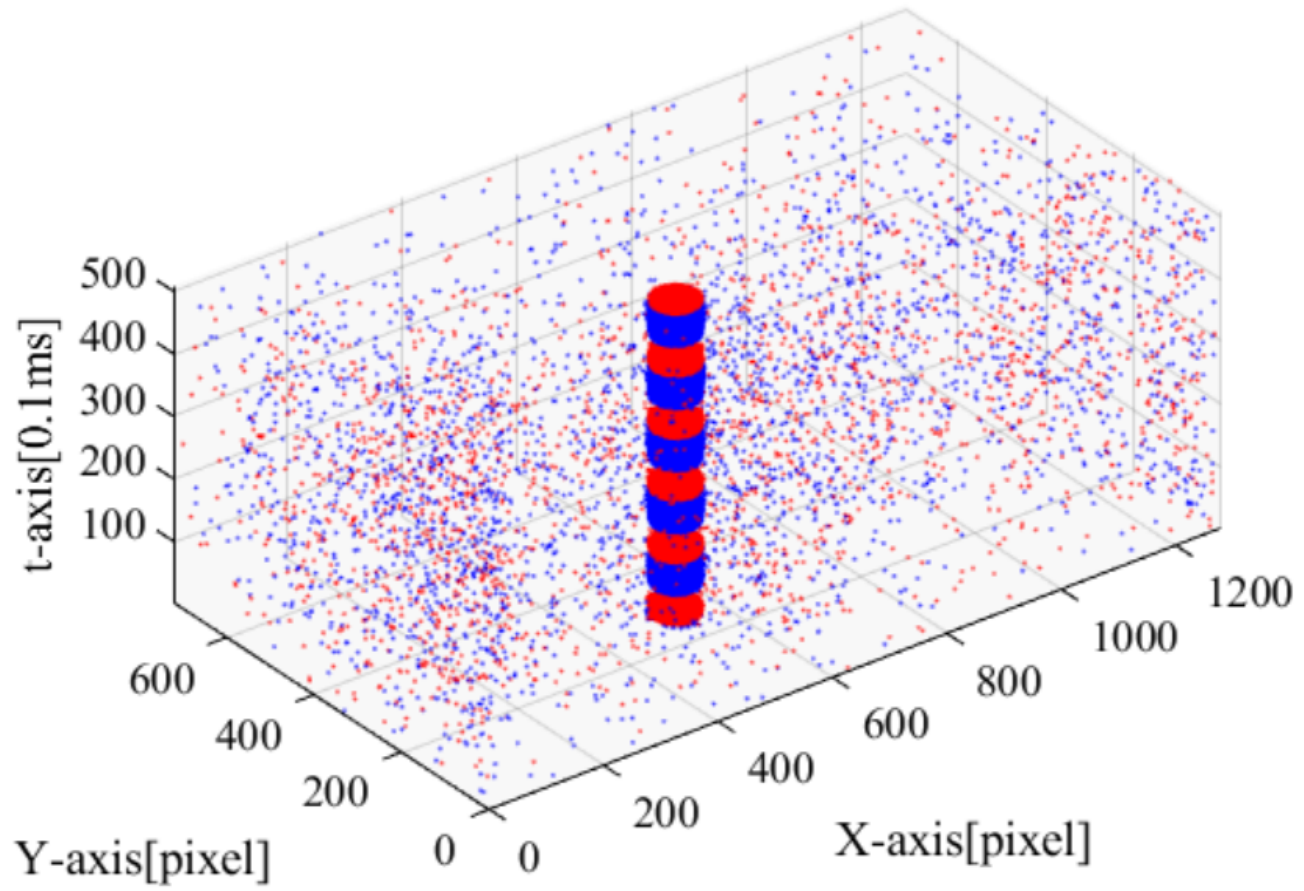}
\end{minipage}
\label{fig7b}
}
\subfigure[Final denoising result by our method.]{   
\begin{minipage}{0.5\linewidth}
\centering 
\includegraphics[height = 4.5cm]{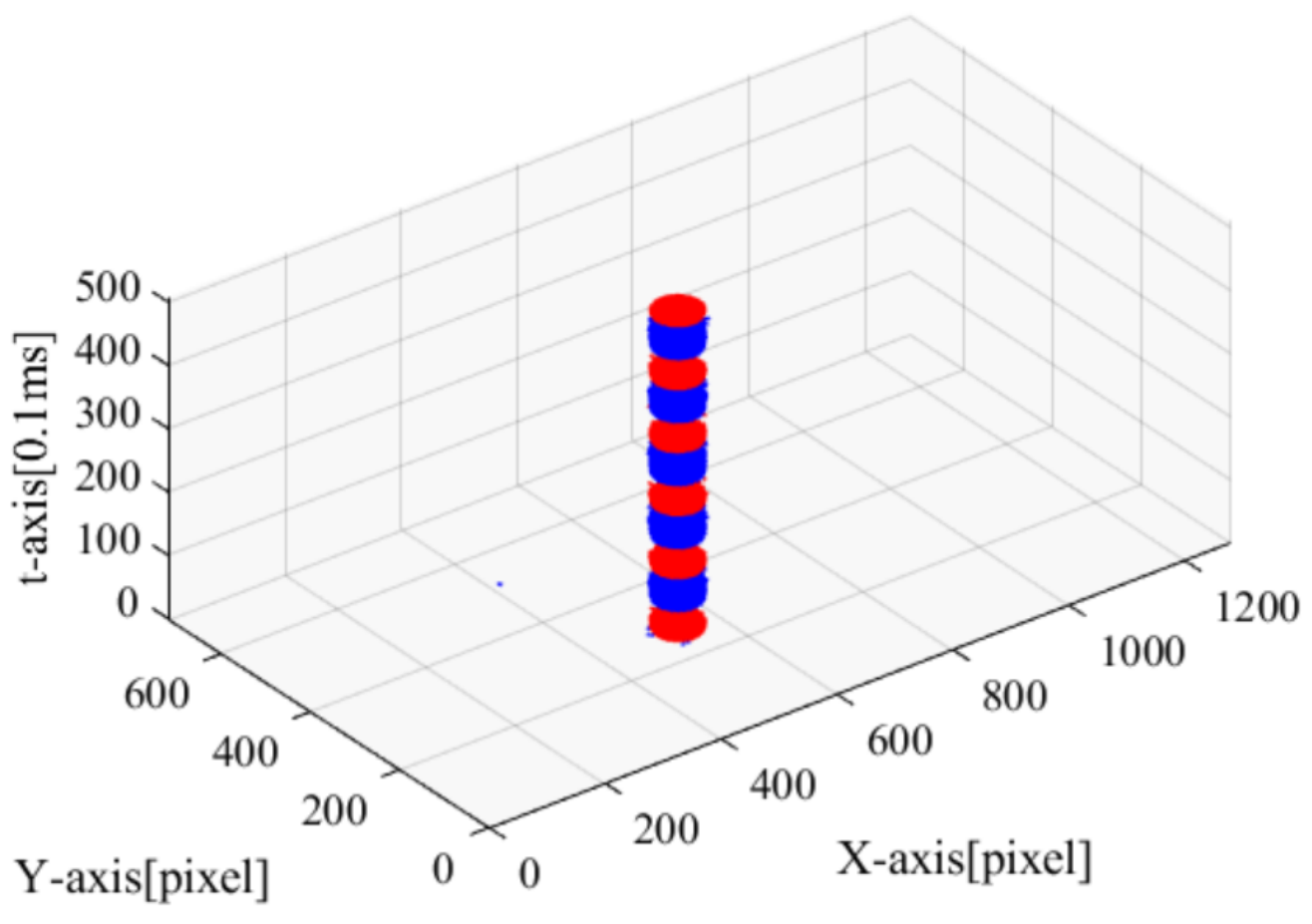}  
\end{minipage}
\label{fig7c}
}\subfigure[Denoising result only by spatiotemporal correlation.]{ 
\begin{minipage}{0.5\linewidth}
\centering  
\includegraphics[height = 4.5cm]{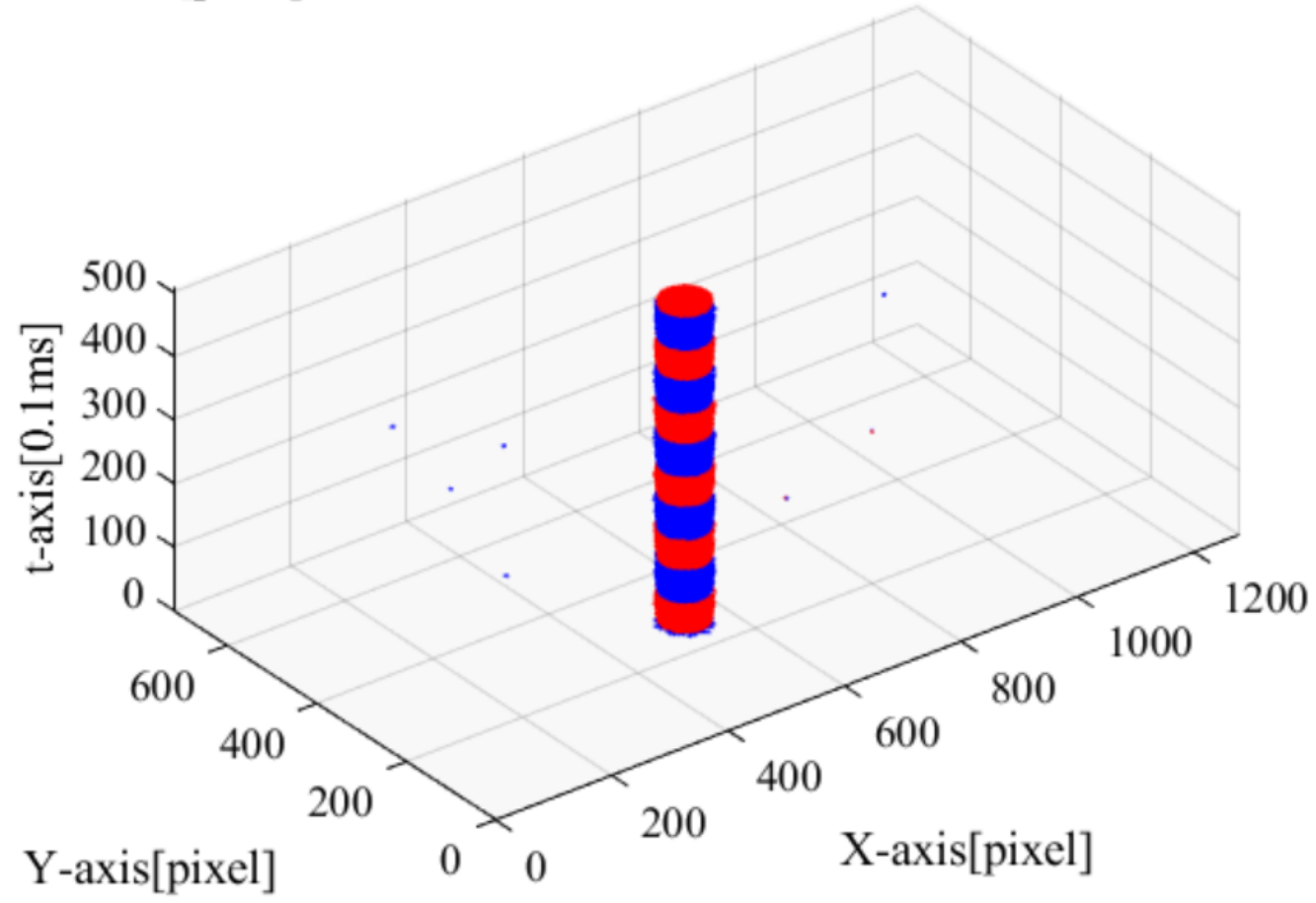}
\end{minipage}
\label{fig7d}
}
\caption{Observation noise suppression.}   
\label{fig7}   
\end{figure}

\begin{table}[htbp]
\centering
\caption{\bf Number of filtered events.}
\begin{tabular}{ccc}
\hline
Method & Number of events & Event removal rate \\
\hline
Original data & 300000 & \\
Spatiotemporal filtering & 272583 & 9.14\% \\
Our method & 217784 & 27.40\% \\
\hline
\end{tabular}
  \label{tab1}
\end{table}

Table. \ref{tab1} presents the number of events after denoising. For the raw event stream containing 300,000 events, applying spatiotemporal correlation filtering alone leaves 272,583 events remaining. In contrast, employing the method proposed in this paper yields 217,784 remaining events, achieving an event removal rate of 27.40\%. The two-step filtering method based on the number of events and spatiotemporal correlation, effectively eliminates observation noise, facilitating the subsequent extraction of the marker center point. However, setting the threshold too high for the number of events during coarse filtration may produce events caused by LED marker blinking, potentially leading to biased extraction of the marker center point.

\subsection{Marker center point extraction}
\begin{figure}[ht]
\centering
\subfigure[LED markers fixed on a rigid, straight steel pole.]{   
\begin{minipage}{0.5\linewidth}
\centering 
\includegraphics[height = 4cm]{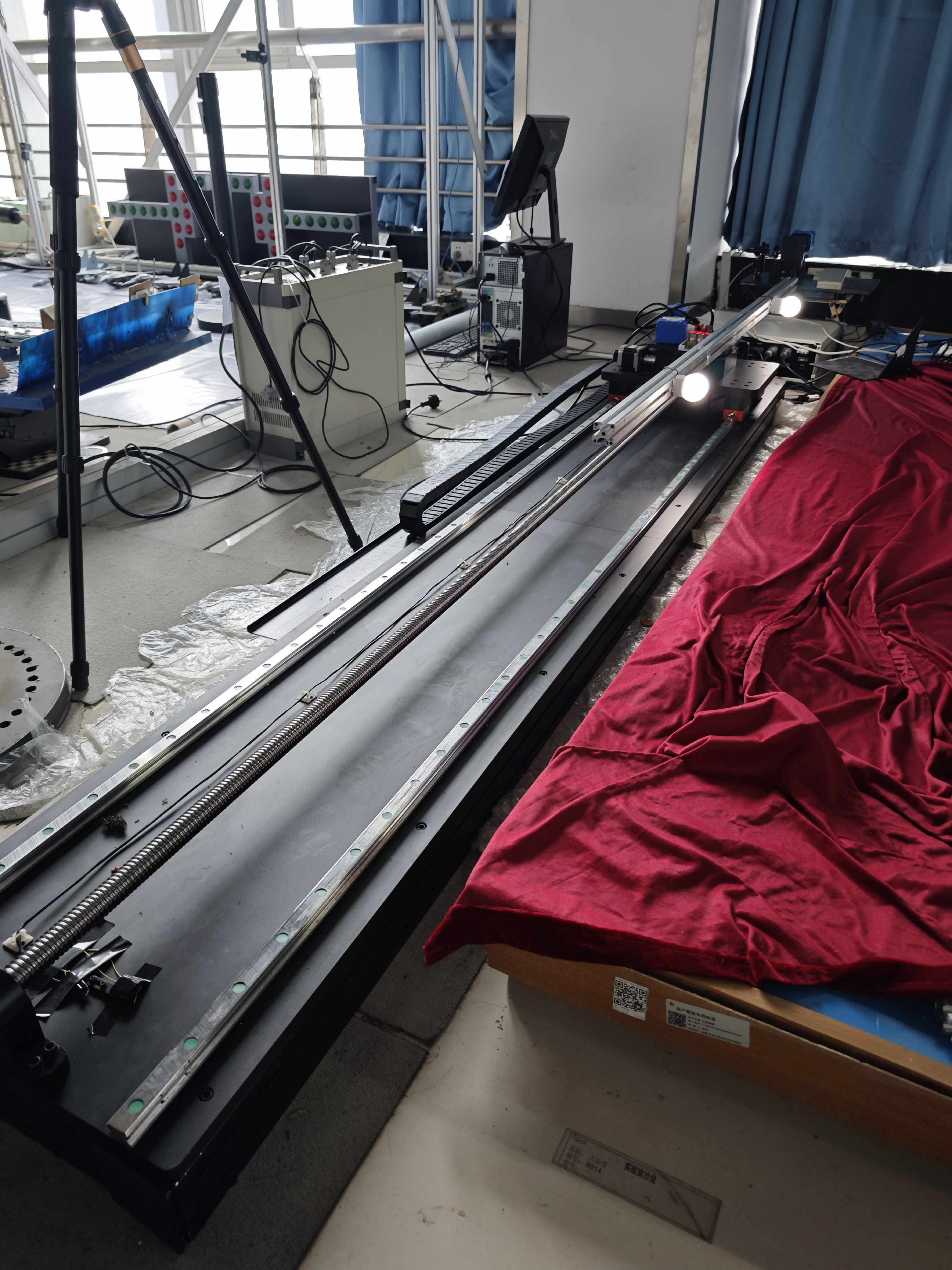}  
\end{minipage}
\label{fig8a}
}\subfigure[Event camera observing LED markers blinking.]{ 
\begin{minipage}{0.5\linewidth}
\centering  
\includegraphics[height = 4cm]{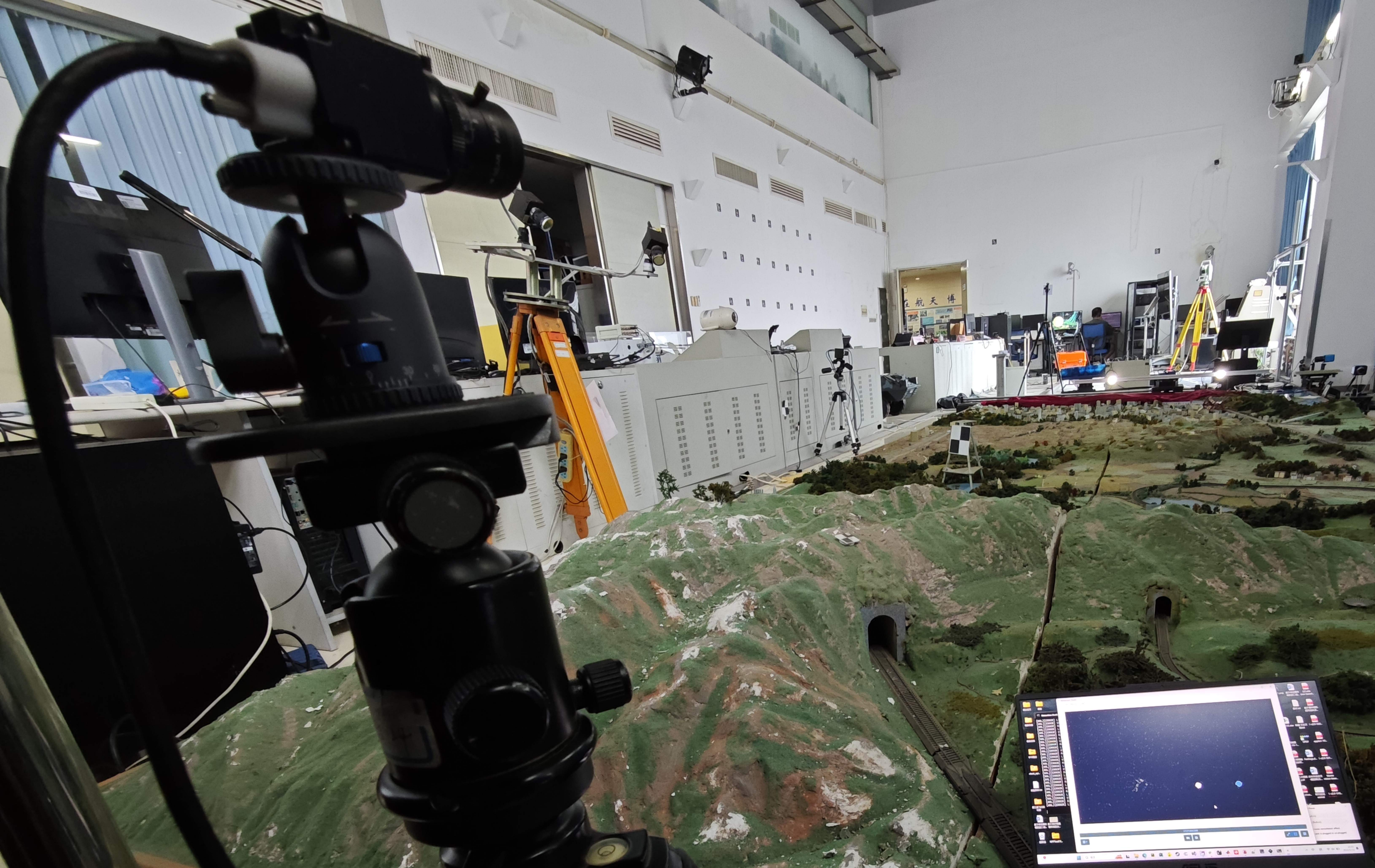}
\end{minipage}
\label{fig8b}
}
\caption{Configuration of the marker extraction experiment.}   
\label{fig8}   
\end{figure}

\begin{figure}[htbp]
\centering  
\includegraphics[height = 4cm]{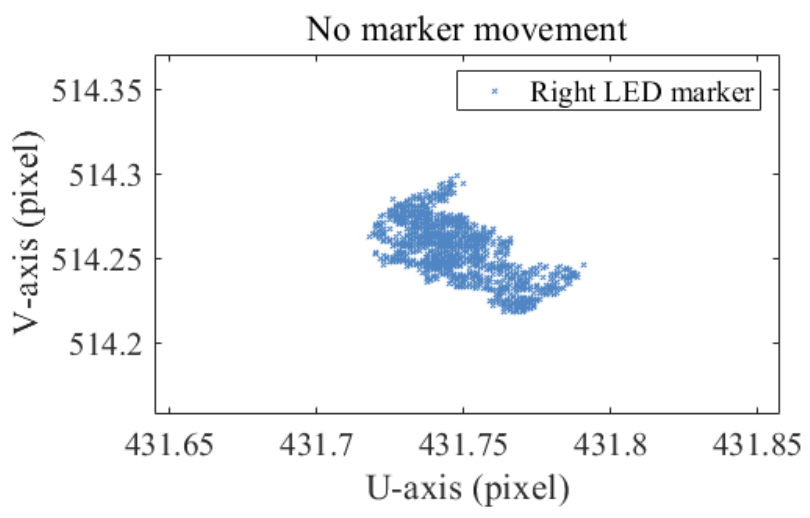}
\caption{Extraction of the marker center point while stationary.}   
\label{fig9}   
\end{figure}

To verify the accuracy of the center point extraction method, we conducted experiments as illustrated in Fig. \ref{fig8}. Due to the challenge of accurately determining the true pixel coordinates of the LED markers, we employ a controlled platform to displace the markers. Subsequently, we measure the displacement in the image to validate the accuracy of extracting the marker center point. Fig. \ref{fig8a} depicts a platform capable of finely controlling displacement movements with exceptional accuracy, achieving adjustable displacements as $1\enspace \mu \text{m}$. Two LED markers, each operating at a blinking frequency of $100\enspace \text{Hz}$, are fixed on a rigid, straight steel pole. The center points of the two markers are separated by a distance of $1\enspace \text{m}$. Then the markers are affixed onto the platform for accurate displacement and use our method for measurement. The optical axis of the event camera is perpendicular to the platform to be tested, and the camera is $8 \enspace\text{m}$ away from the markers, as depicted in Fig. \ref{fig8b}.

At the beginning of the experiment, the markers are first made to be stationary. The accuracy of extracting the marker center points is first verified. As shown in Fig. \ref{fig9}, the maximum change of the marker center point in $U-axis$ is 0.073 pixels and 0.081 pixels in $V-axis$.

Then the markers are displaced to the left by $50\enspace \text{mm}$, $100\enspace \text{mm}$, $500\enspace \text{mm}$, and $800\enspace \text{mm}$ respectively. This process yields the true values of the marker movements. Subsequently, the measurement results are compared against true values to validate the accuracy. Fig. \ref{fig10} illustrates the measurement results of markers movement. Fig. \ref{fig10a} depicts that the right marker shifts by 2.58 pixels and the left marker by 2.71 pixels in response to the marker undergoing an actual displacement of $50\enspace \text{mm}$. When markers move $100\enspace \text{mm}$, the left marker is displaced by 23.35 pixels, and the right marker is displaced by 22.90 pixels, as is seen in Fig. \ref{fig10b}. The error in the displacement of the two center points on the image is caused by the halo LED markers blinking. Nonetheless, this error is less than 0.5 pixels, thereby fulfilling the precision criteria for sub-pixel extraction. Fig. \ref{fig10c} and Fig. \ref{fig10d} show the center point displacement as the markers move $500\enspace \text{mm}$ and $800\enspace \text{mm}$ respectively. 

\begin{figure}[ht]
\centering
\subfigure[Marker moves $50\enspace \text{mm}$.]{   
\begin{minipage}{0.5\linewidth}
\centering 
\includegraphics[width = 6.2cm]{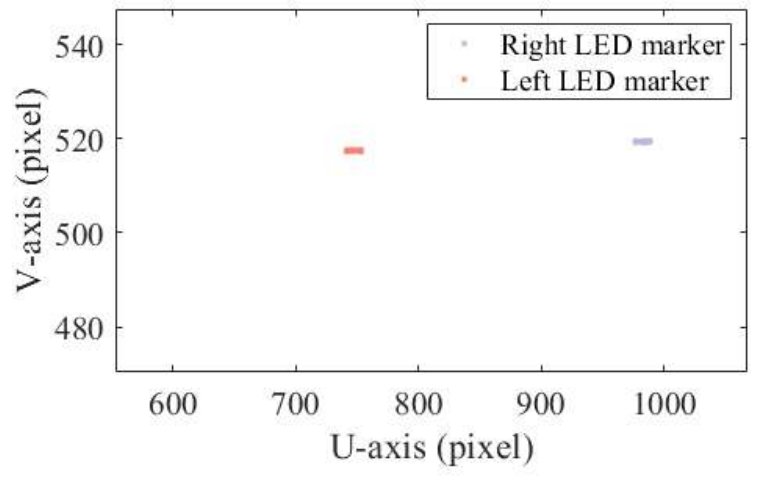}  
\end{minipage}
\label{fig10a}
}\subfigure[Marker moves $100\enspace \text{mm}$.]{ 
\begin{minipage}{0.5\linewidth}
\centering  
\includegraphics[width = 6.2cm]{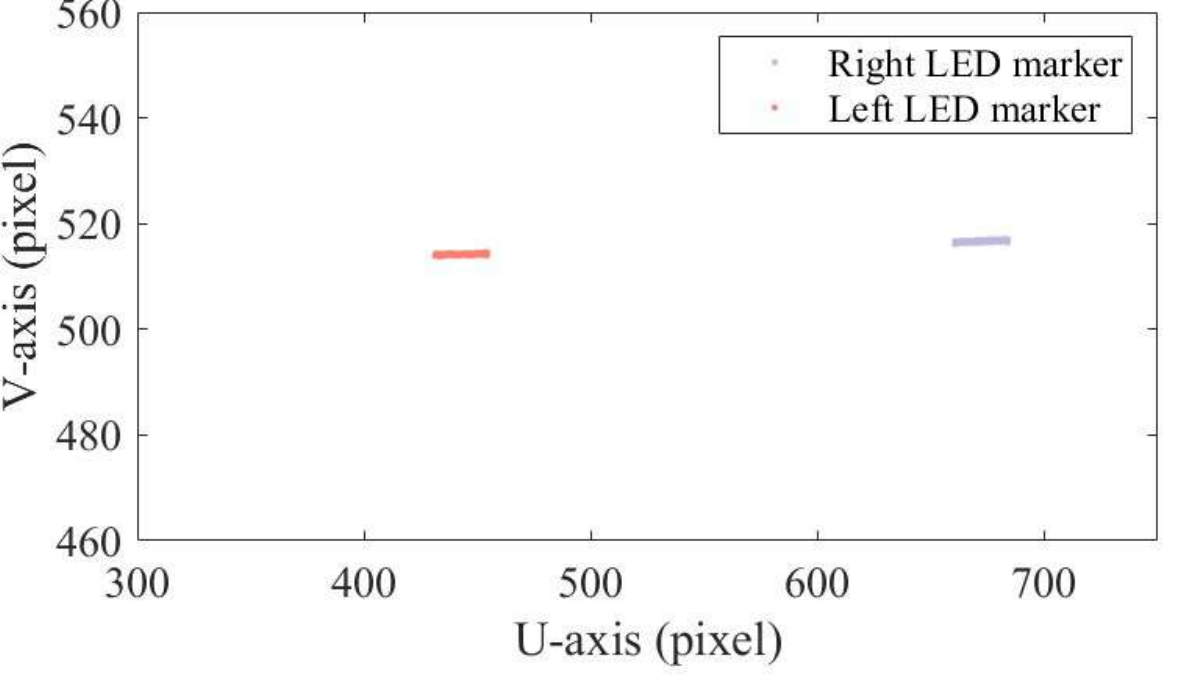}
\end{minipage}
\label{fig10b}
}
\subfigure[Marker moves $500\enspace \text{mm}$.]{   
\begin{minipage}{0.5\linewidth}
\centering 
\includegraphics[width = 6.3cm]{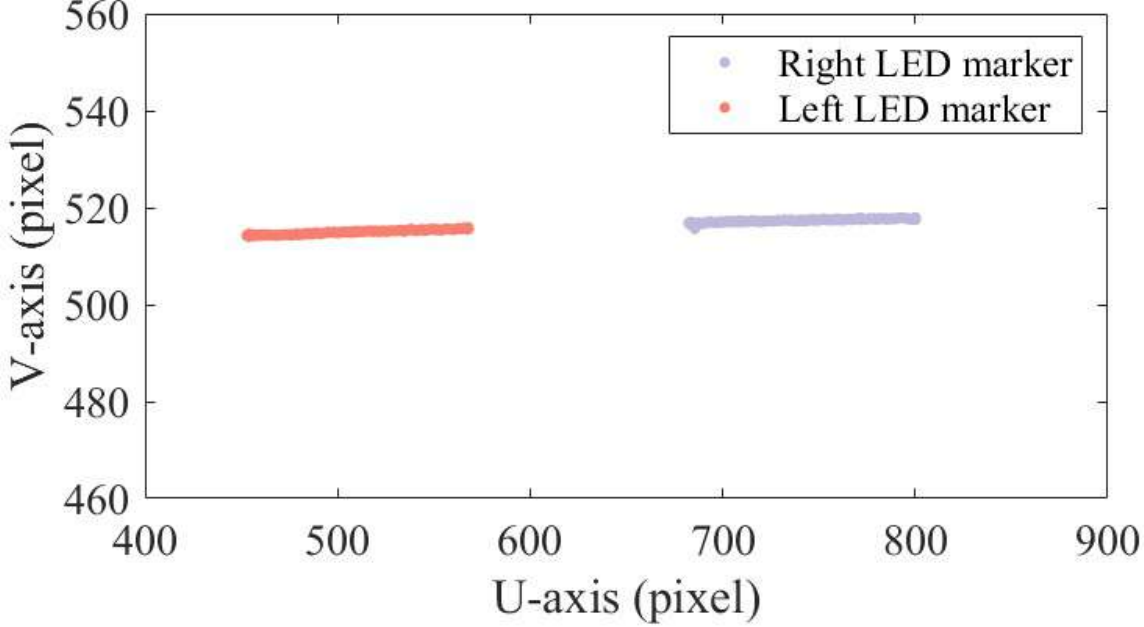}  
\end{minipage}
\label{fig10c}
}\subfigure[Marker moves $800\enspace \text{mm}$.]{ 
\begin{minipage}{0.5\linewidth}
\centering  
\includegraphics[width = 6.5cm]{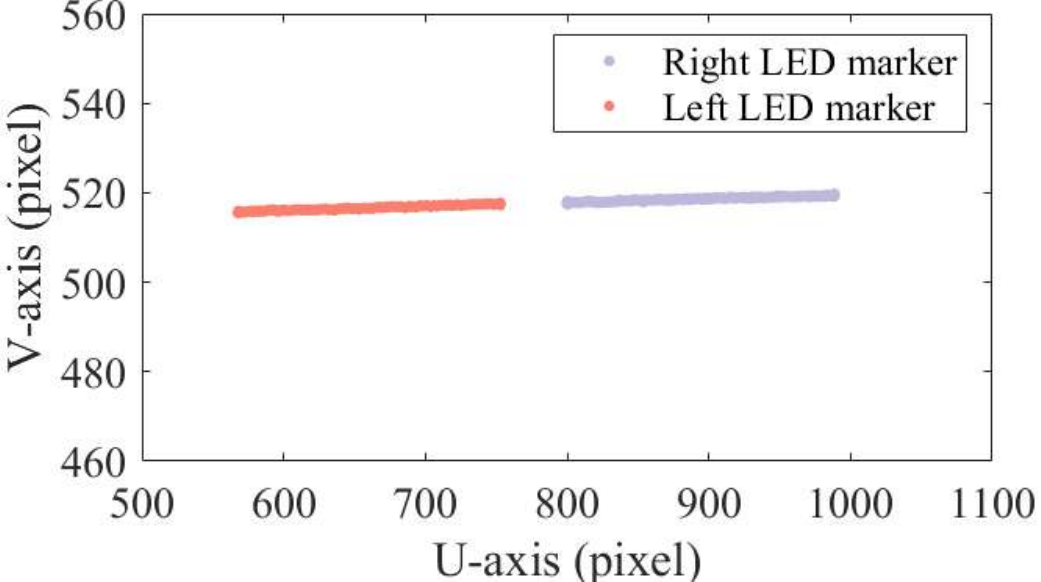}
\end{minipage}
\label{fig10d}
}
\caption{Verify the accuracy of extracting marker center points.}   
\label{fig10}   
\end{figure}

\begin{table}
\centering
\caption{\bf Results of deformation measurement.}
    \begin{tabular}{cccc}
    \hline
        Set displacement distance (mm) & Measurement results (mm) & Measurement error (mm) \\
    \hline
        50  & 50.41  & 0.41  \\
        100 & 100.83 & 0.83  \\
        500 & 499.42 & 0.58  \\
        800 & 800.44 & 0.44  \\
    \hline
    \end{tabular}
    \label{tab2}
\end{table}

Table. \ref{tab2} shows the measurement results. The measurement error of different displacement distances of markers is less than $1\enspace \text{mm}$. When the displacement is under $500\enspace \text{mm}$, it is evident that the measurement error rises as the displacement diminishes. This is attributed to the $500\enspace \text{mm}$ diameter of the LED marker. When displacement is less than this diameter, the LED blink halo introduces events that impact the accuracy of marker center extraction. Experiments demonstrate the accuracy of the marker extraction method proposed in this paper. 

\subsection{High-frequency planar deformation measurement}
\begin{figure}[htbp]
\centering  
\includegraphics[height = 3.5cm]{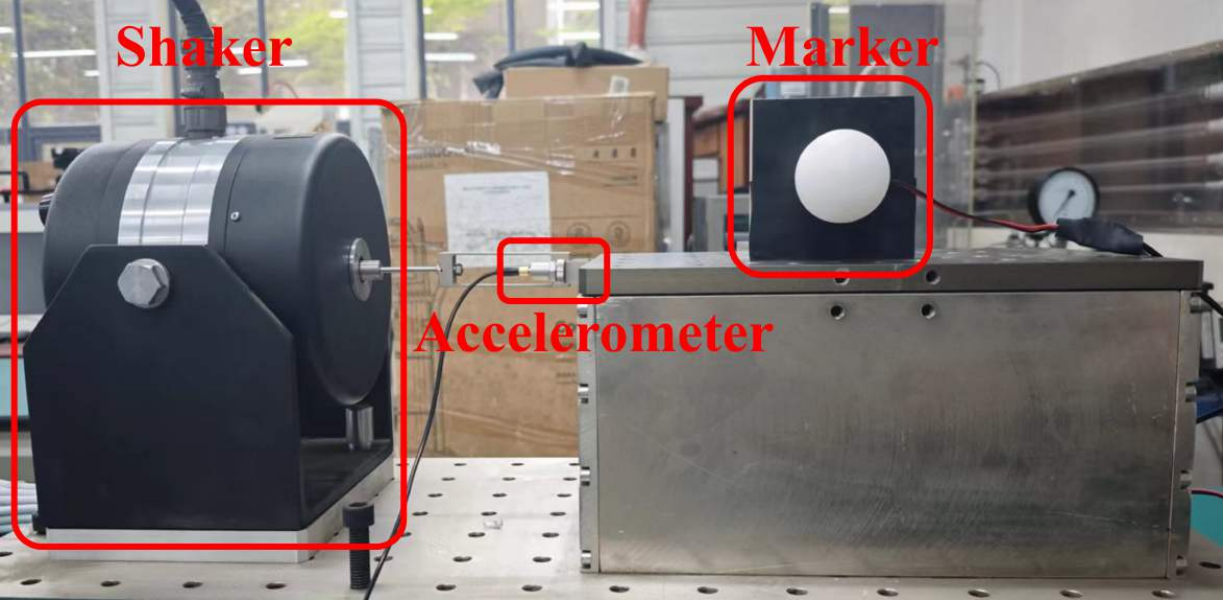}
\caption{Experimental configuration of shaker, accelerometer, and LED marker.} 
\label{fig11}   
\end{figure}

We carried out experiments on high-frequency planar deformation measurements, as depicted in Fig. \ref{fig11}. The experiment employed a shaker capable of vibrating at a specified frequency and waveform. This shaker propelled the platform, generating high-frequency deformation in the plane. The platform was equipped with both an accelerometer and an LED marker. The accelerometer was taken to capture the planar displacement acceleration of the platform, which was then quadratically integrated. Thus the deformation of the platform was obtained. Measurement results by the accelerometer are utilized as true values for high-frequency deformations. 

\begin{figure}[ht]
\centering
\subfigure[LED marker center point extraction.]{ 
\centering  
\includegraphics[width = 9cm]{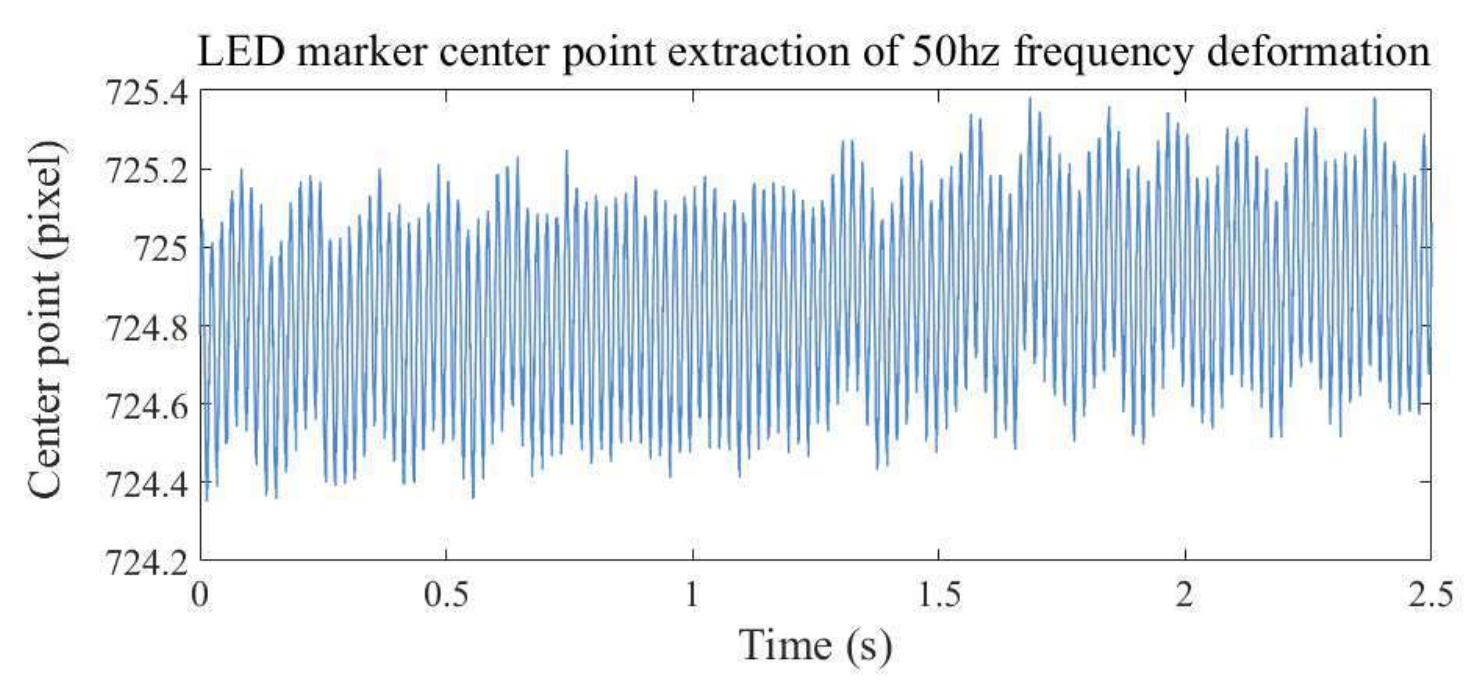}
\label{fig12a}
}
\subfigure[Measurement results.]{   
\centering 
\includegraphics[width = 9cm]{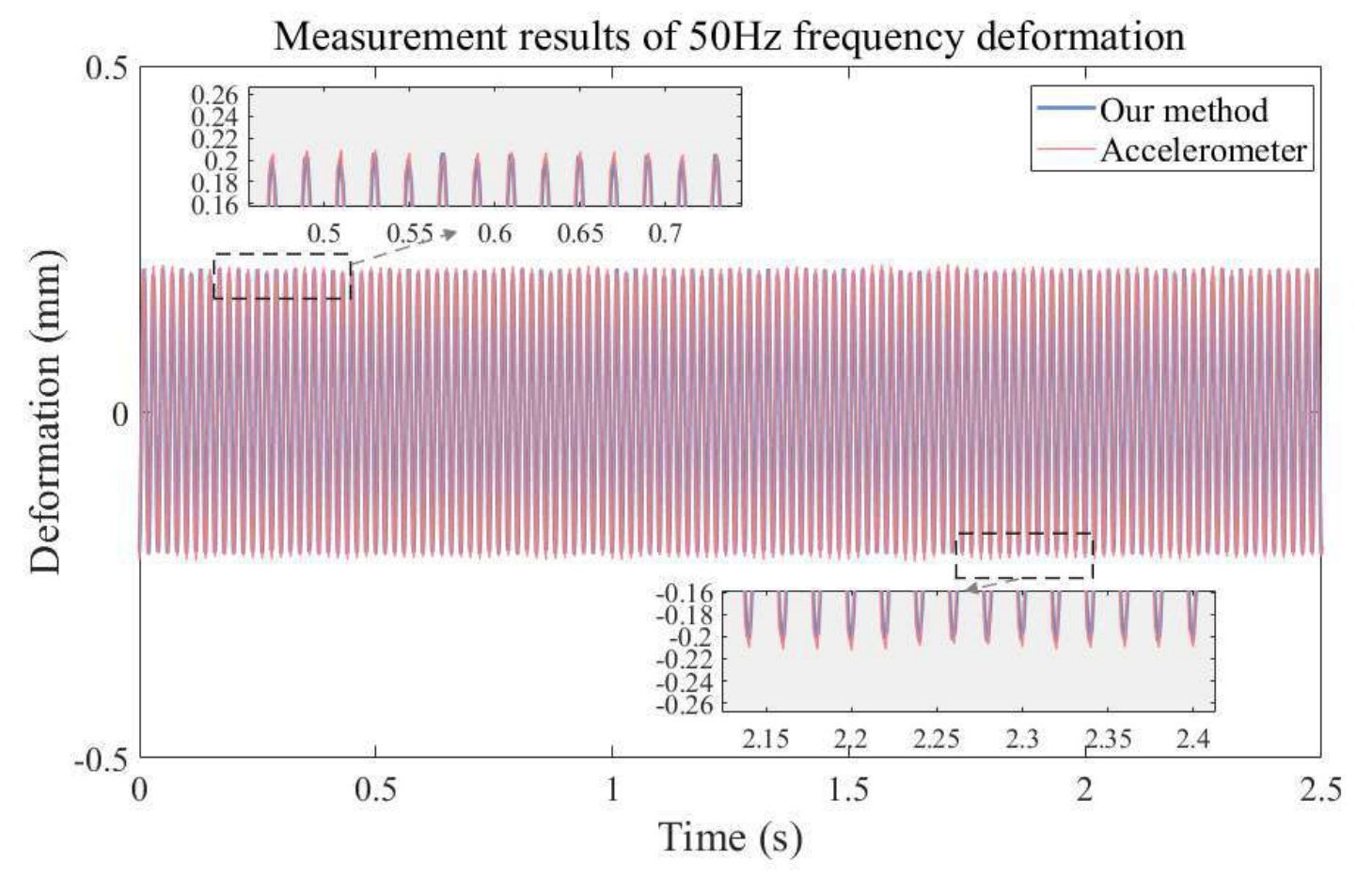}  
\label{fig12b}
}
\caption{Measurement results of high-frequency planar deformation.}   
\label{fig12}   
\end{figure}

A sinusoidal signal of $50\enspace \text{Hz}$ was applied using a shaker, and we measured platform deformations for $2.5\enspace \text{s}$. Fig. \ref{fig12a} shows the extraction results of the LED marker center point. The maximum displacement of the marker moving in the image plane is 0.69 pixels. With the complex external forces acting on the entire experimental platform, low-frequency noise is generated, causing fluctuations in the accuracy of extracting the center of the LED markers. Therefore, a high-pass filter was applied to eliminate low-frequency noise under $1\enspace \text{Hz}$ caused by external forces. The measurement results are shown in Fig. \ref{fig12b}. The measurement results of our method revealed sinusoidal vibrations, with the platform oscillating 125 times within $2.5\enspace \text{s}$, the same as the accelerometer. For the resulting curve of our method, the mean value is $0\enspace \text{m}$, the range is $0.405\enspace \text{m}$, and the standard deviation is $0.144$. Relatively, for accelerometer measurements, the mean value is $0\enspace \text{m}$, the range is $0.41\enspace \text{m}$, as well as the standard deviation is $0.155$. The high-frequency deformation measured by our method is highly fitted to the accelerometer.

The experiment confirms that the method presented in this paper accurately measures high-frequency planar deformation. Under non-contact conditions, the measurement accuracy is comparable to that of the accelerometer.

\section{Conclusion}
This paper proposes a new method utilizing an event camera alongside LED markers to measure high-frequency planar deformations. The event camera offers advantages including high temporal resolution, high dynamic range, low power consumption, and minimal latency. First of all, the noise is filtered based on the characteristics of the event stream generated by LED markers blinking and spatiotemporal correlation, attaining better denoising outcomes. Moreover, the frequency of event polarity reversals is leveraged to differentiate between motion-induced events and events from LED blinking, and the center points of LED markers are extracted by the event cluster, which enables extracting high-speed moving LED markers. Finally, the method measures the high-frequency structure deformation employing a monocular event camera. Through the experiments, the accuracy of the proposed method has been verified.

\section*{Funding}
Hunan Provincial Natural Science Foundation for Excellent Young Scholars (Grant 2023JJ20045); National Natural Science Foundation of China (Grant 12372189).

\section*{Disclosures}
The authors declare no conflicts of interest.

\section*{Data availability}
The data generated and analyzed in the presented research may be obtained from the authors upon reasonable request.

\bibliography{sample}

\end{document}